\documentclass[lettersize,journal]{IEEEtran}
\usepackage{amsmath,amsfonts}
\usepackage{algorithm}
\usepackage{array}
\usepackage{textcomp}
\usepackage{stfloats}
\usepackage{url}
\usepackage{verbatim}
\usepackage{graphicx}
\usepackage{cite}
\usepackage{subfigure}
\usepackage{booktabs}
\usepackage[table]{xcolor}
\usepackage{multirow}
\usepackage{amssymb}
\usepackage[justification=centering]{caption} 

\usepackage{algpseudocode}   

\begin{document}

\title{Homogeneous Low-Resolution Face Recognition Method based Correlation Features}

\author{Xuan Zhao,~\IEEEmembership{Yanching Institute of Technology}
}

%

\maketitle
\begin{abstract}
Face recognition technology has been widely adopted in many mission-critical scenarios like means of human identification, controlled admission, and mobile device access, etc. Security surveillance is a typical scenario of face recognition technology. Because the low-resolution feature of surveillance video and images makes it difficult for high-resolution face recognition algorithms to extract effective feature information, Algorithms applied to high-resolution face recognition are difficult to migrate directly to low-resolution situations. As face recognition in security surveillance becomes more important in the era of dense urbanization, it is essential to develop algorithms that are able to provide satisfactory performance in processing the video frames generated by low-resolution surveillance cameras. This paper study on the Correlation Features-based Face Recognition (CoFFaR) method which using for homogeneous low-resolution surveillance videos, the theory, experimental details, and experimental results are elaborated in detail. The experimental results validate the effectiveness of the correlation features method that improves the accuracy of homogeneous face recognition in surveillance security scenarios.

\end{abstract}


\section{Introduction}
\label{sect:intro}  

Image processing technologies and applications, such as face detection, recognition, and retrieval in surveillance video, play an increasingly important role in today’s security scenario. For example, in the public safety field, many monitoring devices have been deployed in public areas with high population density, enabling fully automatic identification and retrieval of targets and their motion collection in the video. Face recognition at low resolution is still a problem to be solved. The process of face recognition in low resolution is described in article\cite{nagothu2020smart} with detail. The face recognition process in surveillance video can be divided into two parts: Preprocessing and Recognition. In this article, we discuss the method of feature extraction in the face recognition process.

In the paper\cite{luevano2021study}, pictures with a pixel area smaller than 32 × 32 pixels are defined to very low resolution (VLR). Luis et al.\cite{luevano2021study} categorized face recognition applications into homogeneous and heterogeneous. Heterogeneous face recognition matches images from different domains: the probe's very low resolution (VLR) photos and high-resolution gallery images. Relatively, homogeneous face recognition is to match images that come from the same source domain. In the case, we discussed here, both the probe and gallery images come from the low-resolution domain. This paper\cite{luevano2021study} summarizes the super-resolution reconstruction method for Heterogeneous low-resolution face recognition and further subdivides it into Projection methods and Synthesis methods. In addition, it also discusses the effectiveness of lightweight neural networks in homogeneous low-resolution face recognition. 

Similarly to paper\cite{luevano2021study}, Li et al.\cite{li2018face} raised the issue of Low-quality face recognition (LQFR) and classify it into two scenarios: Watch-list identification and Re-identification. In this article, The authors summarized the low-resolution face recognition method into four schemes: Super-resolution reconstruction, Low-quality robust feature, Unified Space, and Deblurring\footnote{The author in the paper\cite{li2018face} also mentioned the de-blurring way. The image quality degradation caused by suspiciousness does not belong to the scope of the low range. Therefore, de-blurring is not mentioned in this article.}. Super-resolution reconstruction methods are classified into two types: Visual quality-driven face super-resolution and Face super-resolution for recognition. The previous one focuses on reconstructing visually high-quality images, and the latter focuses on super-resolution reconstruction for face recognition. The low-quality robust feature is a method using artificially selecting features. Compared with those methods based on deep learning, they are fast and do not rely on training, but when the texture information captured in the LR face image is less, it will have significant limitations. Hand-made features are also susceptible to posture, light occlusion, and expression changes. The unified Space method focuses on learning similarities in common space mapping from LR probe images and HR gallery images. This article is mainly for the homogenized face recognition scene, also regarded as LR-LR face recognition.

\subsection{Contribution of This Article}

This paper conducts further research on the Correlation Features-based Face Recognition (CoFFaR) method for low-resolution surveillance videos. 
The correlation features data preprocessing methods, which significantly increase the amount of learning data and better improve the specificity of correlation features. Further, This article elaborates on two generating ways for different data sets volume. This paper evaluates the impact on the performance of the model of different data set generating methods. At the same time, it also compares the CoFFaR approach with other models more comprehensively.

The main contributions of this paper are as follows:

\begin{enumerate}
	\item Correlation Features based Face Recognition(CoFFaR) is proposed and mathematically demonstrated.
	\item Proposed two different data preparation modes and conducted comparative experiments.
	\item The method is experimentally verified by analyzing the results of several comparative experiments, which were carried out using the proposed CoFFaR method. 
	
\end{enumerate}

\subsection{Article Structure}

 The remainder of this article is organized as follows. Section \ref{sect:rel} briefly introduces low-resolution face recognition related work based on different technical methods. Section \ref{sect:method} describes the theoretical principle of CoFFaR method in detail. Section \ref{sect:exp} statement on the data set and its processing method, metrics, and experimental results. Section \ref{sect:conclusions} expressed the conclusion of this paper and discussed the future work.
\section{Related Work}
\label{sect:rel}

Zhang et al.\cite{zhang2015coupled} propose coupled marginal discriminant mappings (CMDM) method, which projects the data points in the original high and low resolution features into a unified space to achieve classification. Biswas et al.\cite{biswas2011multidimensional}\cite{biswas2013pose} propose to use multi-dimensional scaling to simultaneously convert features from poor-quality detection images and high-quality gallery images so that the distance between them is similar to the distance of the probe image captured under the same conditions as the gallery. Zhou et al.\cite{zhou2011low} propose an effective method called Synchronous Discriminant Analysis (SDA), which can learn two mappings from LR and HR images to a common subspace to distinguish attribute maximization. Siena et al.\cite{siena2012coupled} propose a method of learning high-resolution and low-resolution two sets of image projections based on the local relationship in the data. The paper\cite{8303213}\cite{8575553}\cite{8682384} uses a deep learning-based method to implement coupled mapping in different resolution domains. The Deep Coupled ResNet method\cite{8303213} proposed by Lu et al. uses a residual network as the primary network to extract robust features, which is used explicitly for feature extraction across different resolutions. Sivaram et al. introduce different types of constraints in the GenLR-Net\cite{8575553} method so that the method can recognize low-resolution images and generalize well to unseen images. Zha et al. propose a transferable coupling network (TCN)\cite{8682384}, which bridges the domain gap by learning LR subnet parameters from fixed and pre-trained HR subnets and their optimization process. The S2R2\cite{4587810} matching method proposed by Hennings-Yeomans et al. can simultaneously perform super-resolution and face recognition in the low-resolution image.

The methods introduced above all belong to the heterogeneous method. An alternative way is to extract features directly from the face in an end-to-end manner; it is also considered a homogeneous method. Although the authors mentioned end-to-end recognition methods in the paper\cite{luevano2021study}, they only introduced several lightweight convolutional neural networks that can be used for face recognition. They did not mention end-to-end recognition methods for low-resolution face recognition. The paper\cite{zhao2019face}\cite{zhao2019faceproquest} proposed CoFFaR scheme takes the end-to-end approach, using correlation feature extracting from low-resolution face image pairs.

\section{CoFFaR: Correlation Features based Face Recognition}
\label{sect:method}
Section \ref{sect:rel} mentioned the Correlation Features based Face Recognition (CoFFaR) method, which based on correlation for low-resolution face recognition. This paper conducts further research on the CoFFaR method, explaining its theoretical and experimental details in-depth. This section introduces the face recognition method based on correlation relationships in detail. The optimization method is explained from the perspective of information entropy, which is optimized by minimizing the Kullback–Leibler distance of the entropy of the output distribution from the correlation filter and the entropy of the ideal correlation distribution. Finally, the details of the network layers involved in the deep learning-based correlation feature extractor are introduced.

\subsection{Notations}
\label{sec:3.1}

The proposed CoFFaR scheme is described using the following notations:

%
\begin{itemize}
    \item $f$: the input image;
    \item $W$: the correlation filter (Correlation feature extractor);
    \item $g$ and $g(f(x))$: the ideological output of correlation probability distribution;
    \item $g'$ and $g(f(x))$: the output after filtering the input image using the correlation filter;
    \item $w(x, y)$: the convolution kernel;
    \item $f(x, y)$: the input map;
    \item $g'(x, y)$: the $n$-th hidden layer output in filter;
    \item $H(g)$: the entropy of the output probability distribution correlation relationship under ideal conditions;
    \item $H(g)'$: the entropy of the probability of output after correlation filtering; and
    \item $H(g||g')$: the relative entropy.
\end{itemize}
 
\subsection{Correlation Features based Face Recognition}
\label{sec:3.2}

Face features are widely applied for biometric verification. 
At present, the mainstream face recognition methods based on deep learning usually extract the features of images using CNNs, followed by a custom loss function to minimize the distance between certain features such that faces are verified. Image data for low-resolution face recognition is scarce because of the difficulties in accurate feature extraction with few pixels-per-face. Under this situation, it is impossible to obtain a precise face recognition performance by a specific feature distance in a targeted manner. 

\begin{figure*}
	\centering
	\includegraphics[width=\textwidth]{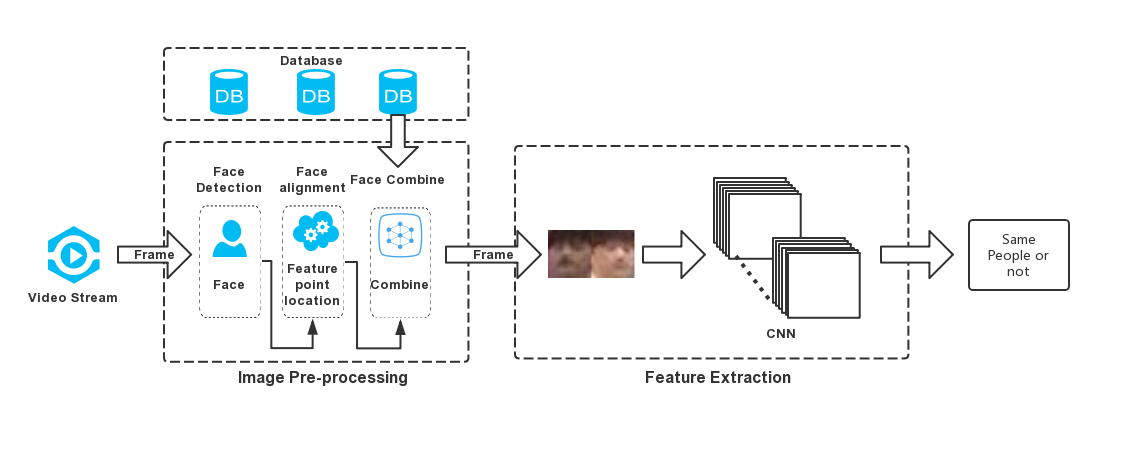}
	\caption{Illustration of the working flow of CoFFaR Scheme.}
	\label{figure:algori}
\end{figure*}

Therefore, this thesis proposes a method based on correlation features. Figure \ref{figure:algori} shows the operational flow of the proposed CoFFaR scheme, in which multiple images are combined and fed the CNNs to extract the features and the correlation among face images rather than being limited to extracted features. In this way, the face information of the low-resolution image can be maximized, and better model performance can be obtained.

\subsection{Theoretical Foundation of CoFFaR Method}
\label{sec:3.3}
Correlation Features based Face Recognition method as shown by the \ref{equ:1}

\begin{equation}
\label{equ:1}
g' = W \cdot f
\end{equation}

\noindent where $f$ is the input image by size of $20 \times 40$, $W$ is the  correlation filter (Correlation feature extractor), $*$ represents the correlation operation, and $g'$ is the output after filtering the input $f$ using the filter $W$.

\begin{equation}
\label{equ:2}
w(x, y) \star f(x, y) = \sum_{s=-a}^a  \sum_{t=-b}^b w(s, t)f(x+s, y+t)
\end{equation}

Equation \ref{equ:2} represents the relevant operations in the image; To facilitate the experiment, we use convolution operations instead of related operations.

\begin{equation}
\label{equ:3}
\begin{aligned}
g'(x, y) &= w(x, y) \ast f(x, y) \\
         &= \sum_{s=-a}^a  \sum_{t=-b}^b w(s, t)f(x-s, y-t)
\end{aligned}
\end{equation}

where $f(x, y)$ is the input map and the $g'(x, y)$ is the output map. w(x, y) is w is a convolution kernel of size $m \times n$, $a = (m - 1)/2 $ and $b = (n - 1) / 2$. The multidimensional vector of the convolutional layer output is flattened and input to the fully connected hierarchy, and then input the vector of the fully connected layer to softmax and get $y_i'$.

\begin{equation}
\label{equ:4}
y_i' = \frac{\exp(g'(x, y))}{\sum_{j = 1}^n \exp(g'_j(x, y))} 
\end{equation}

The final output of $g'(f(x))$ after the input image passes through the filter is the binomial probability distribution output from softmax. When the input image representation has a correlation, which is $f(x) = 0$, then the final output is $y_i$, When the input image representation has no correlation, which is $f(x) = 1$, then the final output is $1 - y_i$. 

\begin{equation}
\label{equ:5}
g'(f(x)) = \begin{cases} 
y'_i \quad \quad \ \ \ f(x) = 0\\
1 - y'_i \quad f(x) = 1 \\
\end{cases}
\end{equation}

We only need to minimize absolute distance between $g(f(x))'$ and $g(f(x))$, as shown in Equation \ref{equ:6}. Where $g(f(x))$ is ideal correlation distribution probability of final output.

\begin{equation}
\label{equ:6}
minimize |g(f(x))' - g(f(x))|
\end{equation}

Ideal correlation distribution probability of final output is shown in equation \ref{equ:7}. When $f(x)$ is the relevant face pair, $g(f(x)) = y_i' = 1$, When $f(x)$ is the irrelevant face pair, $g(f(x)) = 1 - y_i' = 1$.

\begin{equation}
\label{equ:7}
\begin{aligned}
g(f(x)) = \begin{cases} 
y_i 	 \\
1 - y_i  \\
\end{cases}
\end{aligned}
\end{equation}

In summary, by optimizing the filter by minimizing the absolute distance between $g(f(x))'$ and $g(f(x))$, the filter can extract the distribution of image correlation features, and then the predicted probability distribution of the image is finally output.

\subsection{Optimization  Method}
\label{sec:3.4}
The correlation relationship filters the entropy of the output probability distribution after the input image

\begin{equation}
\label{equ:8}
\begin{aligned}
H(g) &= - \sum_{x \in [0, 1]} g(f(x))\log(gf(x)) \\ 
     &=-y_0(x)\log y_0(x) - y_1(x)\log y_1(x)  \\
\end{aligned}
\end{equation}

\noindent where $y_0$ is irrelevant probability and $y_1$ is relevant probability. Since this is a binomial distribution, $y_0 = 1 - y_1$. The Equation \ref{equ:8} can be rewritten as below:

\begin{equation}
\label{equ:9}
\begin{aligned}
	H(g) = &-y_1(x)\log y_1(x) \\ 
		   &-(1 - y_1(x))\log (1 - y_1(x))
\end{aligned}
\end{equation}

Entropy is a measure of the uncertainty of a random variable. $H(g)$ represents the output uncertainty of the correlation under ideal conditions,therefore $H(g) = 0$.

\begin{equation}
\begin{aligned}
	\label{equ:10}
H(g') = &-y'_1(x)\log y'_1(x) \\
	    &-(1 - y'_1(x))\log (1 - y'_1(x))
\end{aligned}
\end{equation}

$H(g')$ is the entropy of the probability of output after correlation filtering. According to Equation \ref{equ:6}, the goal of our method is to minimize absolute distance between $g(f(x))'$ and $g(f(x))$.From Equation \ref{equ:7}, we can see that $g$ is the correlation probability distribution under ideal conditions. So we can minimize the Kullback–Leibler distance between $H(g')$ and $H(g)$, As formula \ref{equ:11}.

\begin{equation}
\label{equ:11}
minimize \ D_{KL}||H(g'), H(g)|| 
\end{equation}

Minimizing $D_{KL}$ is also to minimize the relative entropy between $H(g')$ and $H(g)$. The relative entropy is calculated as shown in Equation \ref{equ:12}.

\begin{equation}
\label{equ:12}
H(g||g') = H(g, g') - H(g) 
\end{equation}

\noindent where the H(g, g') is shown in below,
\begin{equation}
\label{equ:13}
\begin{aligned}
H(g, g') = &-y_1(x)\log y'_1(x) \\
		   &-(1 - y_1(x))\log (1 - y'_1(x)) 
\end{aligned}
\end{equation}

According to the nature of the relevant entropy, $H(g||g') \geq 0$. Based on Equation \ref{equ:7}, $g$ is the correlation probability distribution under ideal conditions, and its uncertainty is 0, so $H(g) = 0$. In order to minimize the relative entropy and make the probability distribution of the correlation filter output closer to the correlation probability distribution in the ideal state, we can only minimize the $H(g, g')$ term, which is cross-entropy.

By minimizing H(g, g'), we minimize the Kullback–Leibler distance of the entropy of the output distribution from the correlation filter and the entropy of the ideal correlation distribution. Because $g$ is the distribution under ideal conditions, minimizing the Kullback–Leibler distance between $g$ and $g'$ means minimizing absolute distance between $g(f(x))'$ and $g(f(x))$. Using this method to train the correlation filter model, we can finally obtain a model whose correlation probability distribution entropy is close to zero. Experiments show that the model has achieved good results for face verification at low resolution.

\subsection{Correlation Feature Extractor Based on Deep Learning}
\label{sec:3.5}

In section \ref{sec:3.4}, a method was introduced to minimize the Kullback-Leiber distance of the entropy of the output distribution from the correlation filter and the entropy of the ideal correlation distribution, then minimize the absolute distance between $g(f(x))'$ and $g(f(x))$. The extraction of correlation features is the most crucial part of face recognition. The method described in this paper uses deep learning methods for doing feature extraction. In the CoFFaR method, convolution operations can be used to replace related operations for feature extraction. The convolution operation process is shown in the formula \ref{equ:14}.

\begin{equation}
\label{equ:14}
y_j = \sum_{i}( k \ast x_i) + b_j
\end{equation}

\noindent where $k$ is the convolution kernel, $x_i$ is the input map, and $y_j$ is the output map, $b_j$
is the bias of the output map. $\ast$ means convolution operation.

\section{Experimental Results}
\label{sect:exp}
\subsection{Dataset}
In this paper, A surveillance face recognition dataset called QMUL-SurvFace\cite{cheng2018surveillance} has been introduced in experiments. In this dataset, 463,507 facial images of 15,573 unique identities have been captured; the low-resolution facial images are not synthesized by artificial down-sampling of high-resolution imagery but are drawn from real surveillance videos.

\begin{figure}
	\centering
	\includegraphics[width=0.45\textwidth]{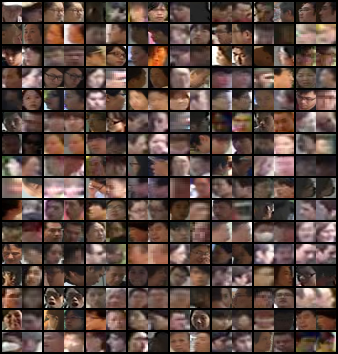}
	\caption{Batch of Imagepair}
	\label{figure:imagepair example}
\end{figure}

The role of image preprocessing in the method proposed in this paper is shown in Figure \ref{figure:algori}. The data used in the CoFFaR method are face pairs. Figure \ref{figure:imagepair example} shows a visualization of the data generated by a batch with a batch size of 128.

\begin{figure}
	\centering
	\includegraphics[width=0.5\textwidth]{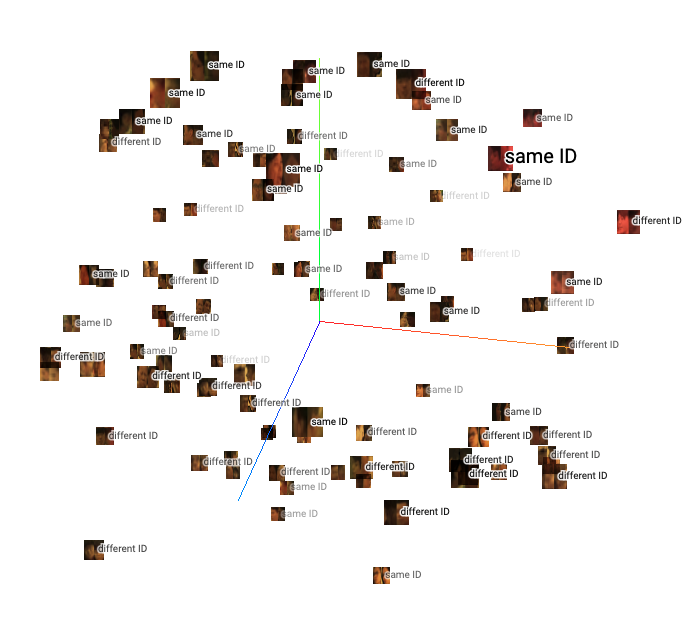}
	\caption{3-dimension Dataset Distribution}
	\label{figure:projector-example}
\end{figure}

Figure \ref{figure:projector-example} illustrate the distribution of data already preprocessed in 3D space. The details of the preprocessed algorithm are presented on \ref{sect:5.2}.

\subsection{Data preprocessing}
\label{sect:5.2}
Because the image preprocessing method proposed in this paper will generate massive amounts of data, and in the case of exhaustive arranged data, the data of the same identity and different identities are asymmetrical. This article considers two kinds of size of dataset experiments. The first one is the "Symmetric" sample dataset. In this dataset, We want to achieve the symmetry of the data collective, so its maximum volume is limited to the same identification data. The second one is "exhaustively arranged" sample data sets. In this dataset, we will not consider the symmetry of the dataset and generate as much data as possible.
\subsubsection{Symmetric sample dataset}

In symmetric preprocessing, the goal is to generate a dataset of equal magnitude for each category. According to equation \ref{equ:18}, After the complete array processing, the amount of data of the same identity is $N_s$, the amount of data of different identities will be far more than the amount of data of the same identity. Therefore, in the generation of symmetric data sets, the data amount of the same identity is chosen as the upper limit.

\begin{figure}
	\centering
	\includegraphics[width=0.45\textwidth]{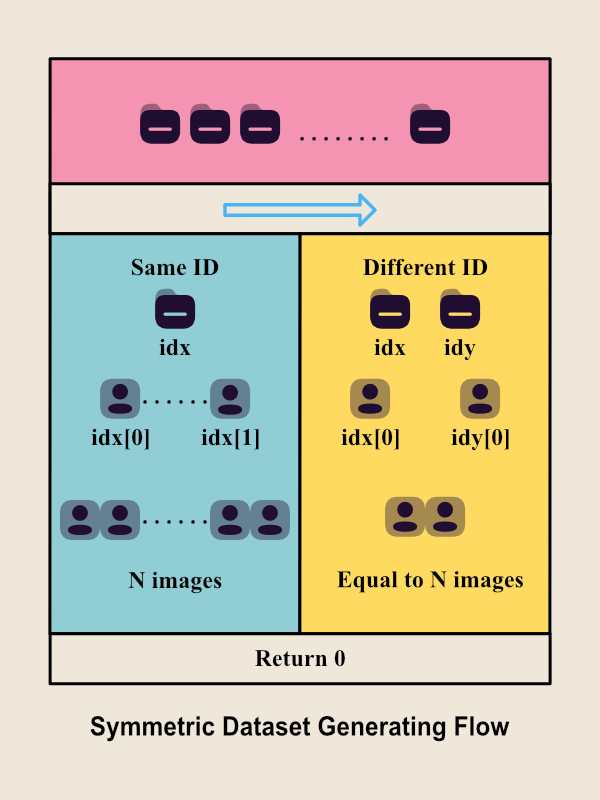}
	\caption{Symmetric dataset generating preprocess algorithm schematic diagram}
	\label{figure:symmetric}
\end{figure}

Algorithm \ref{alg:Framwork-symmetric} is the schematic diagram of the preprocessing flow in symmetric dataset generation. The set of samples in the gallery is presented to $P_G$. The input of Algorithm \ref{alg:Framwork-symmetric} is $P_G$ and output is a dataset that includes the concatenated image. The schematic diagram of symmetric dataset generating preprocess algorithm is as shown in \ref{figure:symmetric}.

There is a total of 220,000 data in the dataset after the entire arrangement after symmetric preprocessing.

\begin{equation}
\label{equ:18}
    N_s = x*(x-1)*n
\end{equation}

\subsubsection{Exhaustively arranged sample dataset}

\begin{figure}
	\centering
	\includegraphics[width=0.45\textwidth]{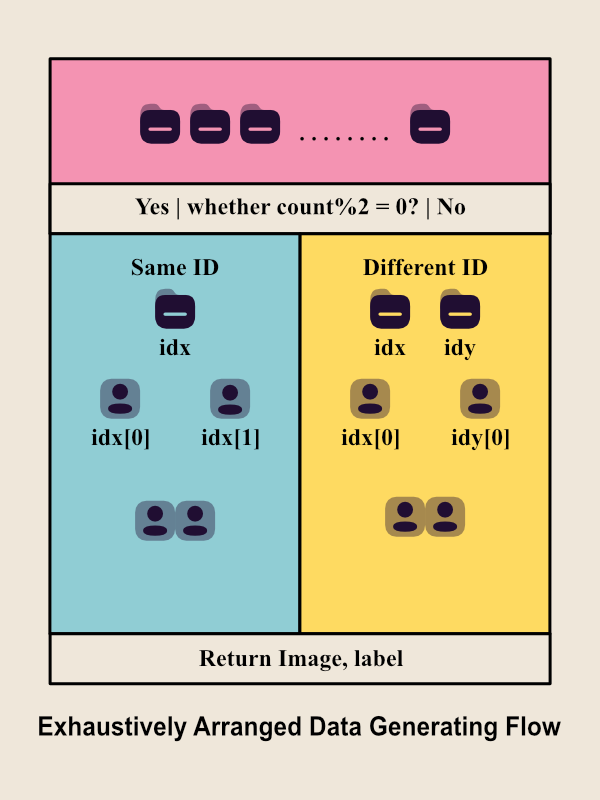}
	\caption{Exhaustively arranged data generating preprocess algorithm schematic diagram}
	\label{figure:exhaustively}
\end{figure}

Algorithm \ref{alg:Framwork-exhaustively} is the preprocessing flow in Exhaustively arranged dataset generation. The set of samples in the gallery is presented to $P_G$. The input of Algorithm \ref{alg:Framwork-symmetric} is $P_G$ and output is  a generator $G_n$. The schematic diagram of exhaustively arranged data generating preprocess algorithm is shown in \ref{figure:exhaustively}.

Theoretically, although the data set produced by this method cannot be called "infinite," it must be said to be massive. Assuming that the average amount of data for each person is $x$, and there are a total of $n$ people, The amount of non-identical data that a single ID can generate is shown in the equation \ref{equ:18}. The amount of non-identical data that all IDs can generate is shown in the equation \ref{equ:19}.

\begin{equation}
\label{equ:18}
    N_d = x^2* (n-1)
\end{equation} 
 
\begin{equation}
\label{equ:19}
\begin{aligned}
    N_a &= N_d * n-1 \\
    	&= x^2 * (n-1)*n 
\end{aligned}
\end{equation}


\begin{algorithm}[H]  
\caption{ Method of data preprocessing - Symmetric dataset generation}  
\label{alg:Framwork-symmetric}  
\begin{algorithmic}[1]  
\Require  
      Samples in gallery which presents as $P_G$; 
\Ensure  
      A $dataset$ of concatenated images;    
\State Acquire $n$ which is how many images the $P_G[id_n]$ included. 
\label{code:preprocessing:acquire}
\State Traversal identifications $P_G[id_n]$ in $P_G$. The traversal will have $n$ steps. Each of step choice a image $P_G[id_n][image]$ and concatenate this image with other images in $P_G[id_n]$ to generate same ID pairs. To do so, each step will generate $n-1$ same-id image pairs.
\label{code:preprocessing:same}
\State When finished the traversal in \ref{code:preprocessing:same}, get the number of how many images in same-id image pairs, denote $N_image$
\label{code:preprocessing:getimagenumber}
\State Traversal in number of $N_image$, The traversal will have $N_image$ steps apparently. Each of step Random choice a ID $P_G[id_1]$, $P_G[id_2]$ from $P_G$, and then random choice a image $P_G[id_1][image]$ in $P_G[id_1]$ and a image $P_G[id_1][image]$ in $P_G[id_1]$ separeatly. Finally concatenate this two images to generate different ID pairs. To do so, each step will generate $1$ image, the traversal will generate $N_images$ different-ID image pairs.
\label{code:preprocessing:different}  
\State Though \ref{code:preprocessing:same} and \ref{code:preprocessing:different}, the dataset should already all-set. 
\label{code:preprocessing:append} \\  
\Return $0$; 
\label{code:preprocessing:return}
\end{algorithmic}  
\end{algorithm}

\begin{algorithm}[H]  
\caption{ Method of data preprocessing - Exhaustively arrange dataset generation}  
\label{alg:Framwork-exhaustively}  
\begin{algorithmic}[1]  
\Require  
      Samples in gallery which presents as $P_G$; 
\Ensure  
      A generator $G_n$;  
\State Initializing random seeds which $count$ equal to 0;  
\label{code:preprocessing:seeds}  
\State If take the remainder of $count$ to 2 is 0, then go to merge a same pair, or go to merge a different pair.
\label{code:preprocessing:mod}  
\State If merge a same pair, random choice a ID $P_G[id]$ from $P_G$,and then random choice two images $P_G[id][image_1]$ and $P_G[id][image_2]$ in $P_G[id]$, finally combine this two images and  goto step \ref{code:preprocessing:append}.
\label{code:preprocessing:same}  
\State If merge a different pair, random choice a ID $P_G[id_1]$, $P_G[id_2]$ from $P_G$, and then random choice a image $P_G[id_n][image]$ in $P_G[id]$ separeatly, finally combine this two images and  goto step \ref{code:preprocessing:append}. 
\label{code:preprocessing:different}  
\State image.append(),goto step \ref{code:preprocessing:return}, return a batch of data.
\label{code:preprocessing:append} \\  
\Return $image, label$; 
\label{code:preprocessing:return}
\end{algorithmic}  
\end{algorithm}

\begin{figure}
	\centering
	\includegraphics[width=0.42\textwidth]{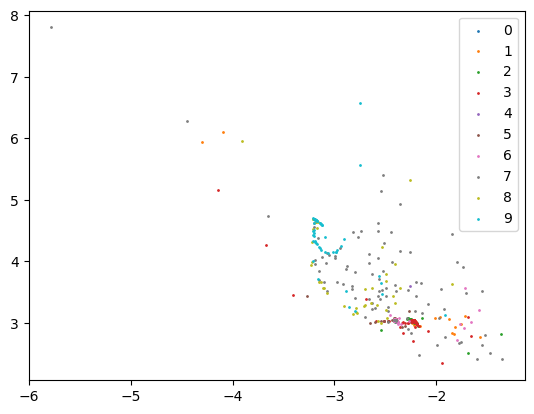}

	\caption{Multi-classification distribution}
	\label{figure:duo-classification distribution}
\end{figure}

\begin{figure}
	\centering
		\subfigure[CoFFaR in training]{
			\includegraphics[width=0.21\textwidth]{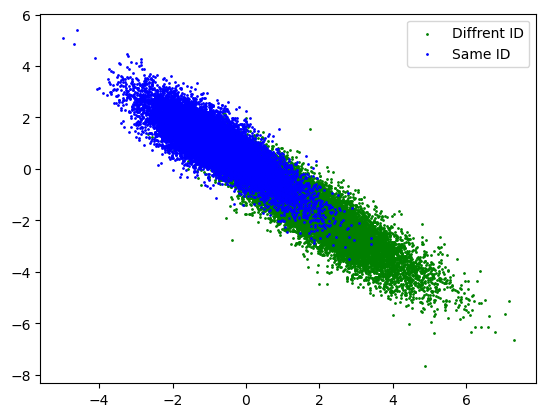}
		}
		\subfigure[CoFFaR in testing]{
			\includegraphics[width=0.21\textwidth]{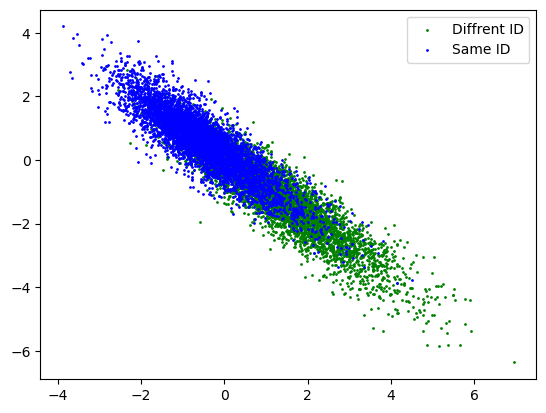}
		}
	\caption{Softmax distribution}
	\label{figure:softmax-distribution}
\end{figure}

\begin{figure}
	\centering
		\subfigure[CoFFaR in training]{
			\includegraphics[width=0.21\textwidth]{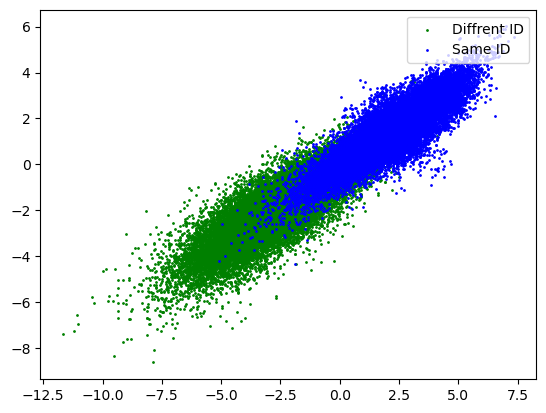}
		}
		\subfigure[CoFFaR in testing]{
			\includegraphics[width=0.21\textwidth]{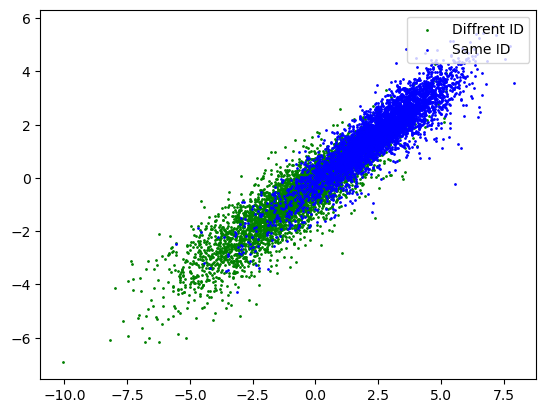}
		}
	\caption{Centerloss distribution}
	\label{figure:Centerloss distribution}
\end{figure}

\begin{table*}
	\centering
	\caption{Performance of Different Algorithms in Face Verification}
	\label{tb:results1}
	\resizebox{\textwidth}{!}{
	\begin{tabular}{c|c|c|c|c|c|c}
	\toprule
  	Methods & TAR@FAR=0.3 & TAR@FAR=0.1 & TAR@FAR=0.01 & TAR@FAR=0.001 &AUC &Accuracy \\
  	\midrule
  	\rowcolor{gray!40} 
  	CoFFaR-softmax & - & - & - & - & - & 82.56\% \\
  	CoFFaR-center & - & - & - & - & - & 77.23\% \\
  	CentreFace & 0.43 & 0.13 & 0.012 & 0.12 & 0.652 & 76.2\% \\
  	DeepID2 & 0.52 & 0.15 & 0.02 & 0.078 & 0.62 & 76.1\% \\
  	FaceNet & 0.49 & 0.163 & 0.293 & 0.012 & 0.529 & 75.3\% \\
  	VggFace & 0.452 &0.16 & 0.201 & 0.04 & 0.85 & 72.1\% \\
  	\bottomrule
	\end{tabular}
	}
\end{table*}

\subsection{Metrics}
\subsubsection{TAR}
TAR(True Accept Rate) represents the ratio of correct acceptance. It means that images with the same identity are predicted to be the same person in the process of face verification. The calculation equation is shown in \ref{equ:19}

\begin{equation}
\label{equ:19}
\begin{aligned}
	P_V(throhold) &= \frac{| \lbrace N:s \geqslant threshold \rbrace|}{P_{same}} \\
	              &= \frac{T_p}{T_p + F_n}
\end{aligned}
\end{equation}

\subsubsection{FAR}
FAR(False Accept Rate) is the ratio of the image be predicted as the same person's image when we compare different people's images.
\begin{equation}
\label{equ:20}
\begin{aligned}
	P_V(throhold) &= \frac{| \lbrace N:s \geqslant threshold \rbrace|}{P_{different}} \\
		          &= \frac{F_p}{F_p + T_n}
\end{aligned}
\end{equation}

\subsubsection{TAR@FAR}
In the case of different FAR values, the value of TAR will also change accordingly.  TAR@FAR represents the value of TAR when FAR takes a fixed value.

\subsubsection{Accuracy}
Accuracy refers to the proportion of all the correct results of the classification model to the total observations.

\begin{equation}
\label{equ:20}
	Accuracy = \frac{T_p + T_n}{T_p + T_n + F_p + F_n} 
\end{equation}

\subsection{Results}

Using this method to train the correlation filter model, we can finally obtain a model whose correlation probability distribution entropy is close to zero. Experiments show that the model has achieved good results for face verification at low resolution.

\begin{figure}
\centering
\includegraphics[width=0.49\textwidth]{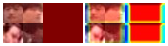}
\caption{A heatmap output from final layer. The activated part of the neuron is evenly distributed on the sample, which further demonstrates that CoFFaR pays more attention to the overall correlation, rather than certain specific features.}
\label{figure:heatmap}
\end{figure}

The distribution of heatmap activated by neurons in the network can further indicate that the CoFFaR method utilizes correlation features. Figure \ref{figure:heatmap} presents the heatmap output from the final network layer. In the heatmap, the activated part of the neuron is evenly distributed on the sample, which further demonstrates that CoFFaR pays more attention to the overall correlation than certain specific features.

Table \ref{tb:results1} illustrate the performance of different algorithms in face verification; the data preprocessing procedure uses an Exhaustively arrange data generating preprocess method. It can be seen from the table that the CoFFaR method has better performance than other methods. CoFFaR-softmax and CoFFaR-center are both experiments based on the correlation method proposed in this paper, but the difference is that they used different loss functions. It is easy to know by its naming that CoFFaR-softmax uses the softmax loss function, and the CoFFaR-centloss uses the center-loss function. The experimental results show that when the CoFFaR method is used, the model performance of the softmax function is better than the center-loss function.

Figure \ref{figure:duo-classification distribution} is feature distribution of multi-category, which is from net of $layer_n$ which using directly classifying method. Figure \ref{figure:softmax-distribution} and Figure \ref{figure:Centerloss distribution} is feature distribution from net of $layer_n$ which using CoFFaR method we proposed in this paper. By comparing Figure \ref{figure:duo-classification distribution} with Figure \ref{figure:softmax-distribution} and Figure \ref{figure:Centerloss distribution}, it can be seen that the dimensionality reduction ability of CoFFaR has a significant effect on the optimization of classification compared to the multi-classification method.
\section{Summary}
\label{sect:conclusions}
\subsection{Conclusions}
This paper conducts further research on the Correlation Features-based Face Recognition (CoFFaR) method of homogeneous face recognition for low-resolution surveillance videos. This paper states that the correlation features data preprocessing significantly increases the volume of learning data and improves the specificity of correlation features. Further, This paper elaborates on two generating ways for different volumes of data sets: the Symmetric Generating Method and the Exhaustively Arranged Method. This paper also evaluates the impact on the performance of the model of different data set generating ways in the CoFFaR approach and compares this approach with other models more comprehensively. Under the low-resolution image dataset, the result using CoFFaR achieved an average accuracy of 82.56\% in the experimental study. 

\subsection{Future Work}
This paper studies the correlation features method applied to the homogeneous scenario in surveillance video. Still, the surveillance system also has a cross-resolution situation, which is also called Heterogeneous. How effective the correlation feature method is applied to heterogeneous face recognition situations remains to be explored. This paper also does not involve the performance evaluation of algorithms in embedded systems, and the embedded systems application is the research direction of future work.

%
%
%
%

\bibliographystyle{abbrv}
\bibliography{ref.bib}


\end{document}